\documentclass[lettersize,journal]{IEEEtran}
\usepackage{amsmath,amsfonts}
\usepackage{algorithmic}
\usepackage{algorithm}
\usepackage{array}
\usepackage[caption=false,font=normalsize,labelfont=sf,textfont=sf]{subfig}
\usepackage{textcomp}
\usepackage{stfloats}
\usepackage{url}
\usepackage{verbatim}
\usepackage{graphicx}
\usepackage{cite}
\usepackage[dvipsnames]{xcolor}

\usepackage{amssymb}
\usepackage{hyperref}
\usepackage[capitalize]{cleveref} 

\newcommand{\myPara}[1]{\vspace{.05in}\noindent\textbf{#1}}

\usepackage{colortbl}
\usepackage[normalem]{ulem}
\useunder{\uline}{\ul}{}

\begin{document}

\title{Temporal-Enhanced Multimodal Transformer for Referring Multi-Object Tracking and Segmentation}

\author{Changcheng~Xiao, Qiong~Cao, Yujie~Zhong, 
Xiang Zhang,~\IEEEmembership{Member,~IEEE}, 
Tao~Wang, Canqun~Yang and Long~Lan,~\IEEEmembership{Member,~IEEE}
\thanks{Changcheng Xiao and Canqun Yang are with College of Computer Science, National University of Defense Technology, Changsha 410073, China (email: xiaocc612@foxmail.com; canqun@nudt.edu.cn).}
\thanks{Long Lan and Xiang Zhang are with Institute for Quantum Information \& State Key Laboratory of High Performance Computing (HPCL), National University of Defense Technology, Changsha 410073, China (email: long.lan@nudt.edu.cn; zhangxiang08@nudt.edu.cn).}
\thanks{Qiong Cao and Tao Wang are with JD Technology, Beijing 102628, China (email: mathqiong2012@gmail.com; bjwangtao@jd.com).}
\thanks{Yujie Zhong is with Meituan Inc., Beijing 100000, China (email: jaszhong@hotmail.com).}
\thanks{\textit{Corresponding author: Qiong Cao.}}

}

\markboth{Journal of \LaTeX\ Class Files,~Vol.~14, No.~8, August~2021}%
{Shell \MakeLowercase{\textit{et al.}}: A Sample Article Using IEEEtran.cls for IEEE Journals}

\IEEEpubid{0000--0000/00\$00.00~\copyright~2021 IEEE}

\maketitle
\begin{abstract}
Referring multi-object tracking (RMOT) is an emerging cross-modal task that aims to locate an arbitrary number of target objects and maintain their identities referred by a language expression in a video. This intricate task involves the reasoning of linguistic and visual modalities, along with the temporal association of target objects. However, the seminal work employs only loose feature fusion and overlooks the utilization of long-term information on tracked objects. In this study, we introduce a compact Transformer-based method, termed TenRMOT. We conduct feature fusion at both encoding and decoding stages to
fully exploit the advantages of Transformer architecture. Specifically, we incrementally perform cross-modal fusion layer-by-layer during the encoding phase. In the decoding phase, we utilize language-guided queries to probe memory features for accurate prediction of the desired objects. Moreover, we introduce a query update module that explicitly leverages temporal prior information of the tracked objects to enhance the consistency of their trajectories. In addition, we introduce a novel task called Referring Multi-Object Tracking and Segmentation (RMOTS) and construct a new dataset named Ref-KITTI Segmentation. Our dataset consists of 18 videos with 818 expressions, and each expression averages 10.7 masks, which poses a greater challenge compared to the typical single mask in most existing referring video segmentation datasets. TenRMOT demonstrates superior performance on both the referring multi-object tracking and the segmentation tasks.
\end{abstract}


\section{Introduction}
\IEEEPARstart{R}{eferring} Multi-Object Tracking (RMOT) is an emerging multi-modal task aimed at the localization of multiple target objects within a video, while simultaneously preserving their identities as specified by a given natural language description. 
In comparison to prior referring tasks  \cite{qiao2020referring, seo2020urvos, khoreva2019video, gavrilyuk2018actor, yu2016modeling, young2014image, kazemzadeh2014referitgame}, RMOT introduces notable challenges \cite{rmot}, particularly revolving around dynamic fluctuations in the number of target objects and pronounced temporal dynamics. To be more precise, multiple objects indicated by the language expression tend to have a more prolonged presence within the video sequence, and the proportion of frames containing these referred objects varies considerably. Hence, RMOT models must possess robust cross-modal comprehension capabilities to precisely identify expected objects and effectively track them in accordance with the query linguistic expressions.

\begin{figure}[!t]
     \centering
      \subfloat[]{\includegraphics[width=\linewidth]{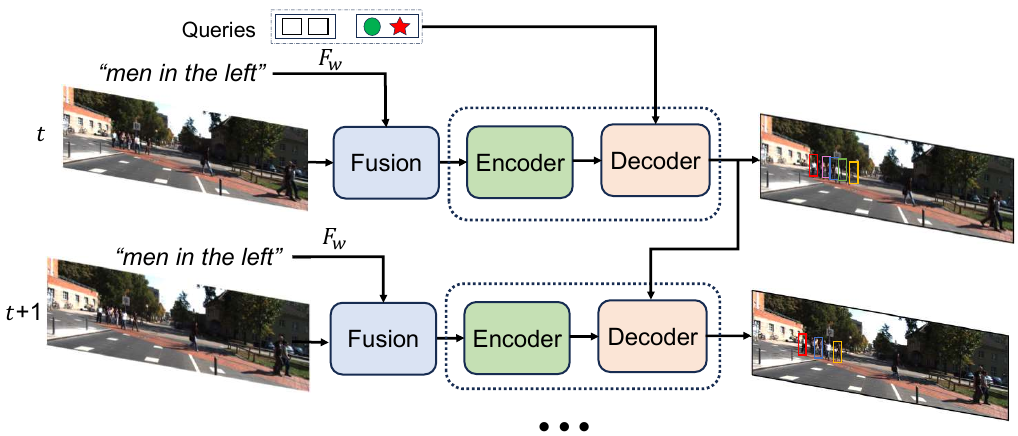}}
         \label{fig: transrmot}
     \vfill
     \subfloat[]{\includegraphics[width=\linewidth]{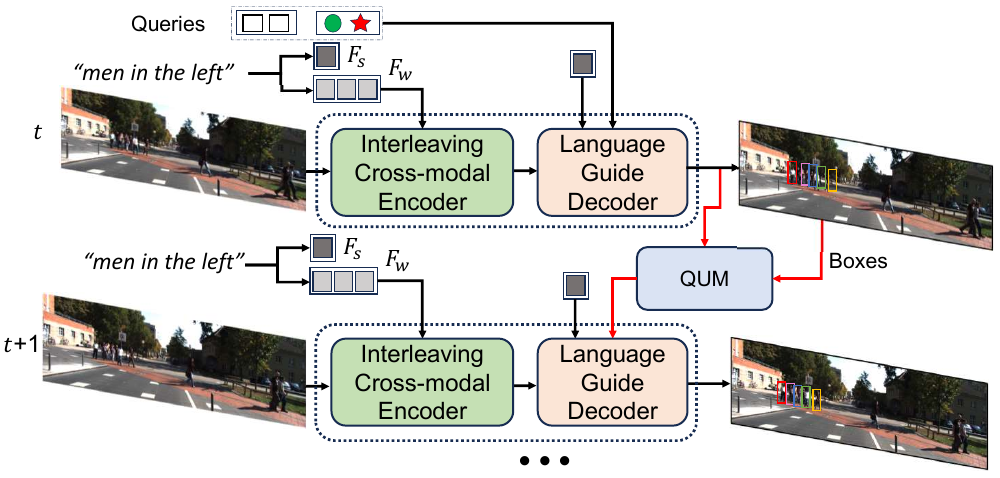}}
         \label{fig: ours}
        \caption{Comparison of the existing referring multi-object tracking pipeline: (a) TransRMOT\cite{rmot} conducts feature interaction solely before the Transformer; (b) TenRMOT, on the other hand, integrates multi-modal feature fusion at both the encoder and decoder. Additionally, TenRMOT capitalizes on the query update module (QUM) to incorporate long-term temporal information regarding the objects. $F_w$ and $F_s$ represent the word-level and sentence-level features of the query expression, respectively. }
        \label{fig:model_comp}
\end{figure}


The pioneering work, TransRMOT \cite{rmot}, is achieved by extending MOTR \cite{motr}, which is based on the encoder-decoder architecture and data association across frames is realized by a continuous update of \textit{track queries}. \IEEEpubidadjcol 
To attain cross-modal inference capabilities, TransRMOT merely integrates word-level features of the referring expression with visual features extracted by the visual encoder through an early-fusion module. Subsequently, this fused information is fed into the encoder for further refinement. This loose feature fusion approach fails to fully leverage the flexible encoder-decoder structure and overlooks the potential guidance of linguistic semantic information during the decoding phase.
Besides, similarly to prior methods  \cite{motr, trackformer}, TransRMOT only relies on short-term information between adjacent frames, which might result in poor performance in scenarios involving occlusions or complex object motion patterns. These oversights may lead to imprecise or inadequate predictions of the intended objects.

To address the above problems, we introduce a new Temporal-ENhanced multimodal transformer for Referring Multi-Object Tracking, termed \textit{TenRMOT}. As shown in \cref{fig:model_comp}, our TenRMOT generally follows the DETR \cite{detr} framework and compactly performs cross-modal feature fusion in both encoding-decoding stages along with a novel query update module for query propagation across frames, realizing accurate tracking of the expected objects.

Firstly, to comprehensively leverage and fuse multi-modal features, we establish interactions between visual and linguistic features at both the encoding and decoding stages. Specifically, in the encoding stage, we introduce an Interleaving Cross-modality Encoder (ICE), which fuses word-level text features and visual features in a layer-by-layer interleaving manner.  This progressive manner enhances the interaction between the visual features and the textual information, allowing the vision-language features generated by the ICE to better highlight the target objects. Besides, in the decoding stage of DETR-based models \cite{detr,defdetr,liu2022dabdetr, wang2022anchor}, the \textit{queries} for the decoder are composed of learned fixed vectors that tend to correspond to specific kinds of modes. To ensure that these queries for the decoder are contextually relevant to the query expression, we inject sentence-level features of the expression into them, guiding their focus toward the referent objects. These strategies collectively lead to a substantial enhancement in grounding the desired objects.

Secondly, to reinforce the consistency of the predicted trajectories, we introduce the inter-frame Query Update Module (QUM), explicitly harnessing temporal information objects to update \textit{track queries}. 
Each query for the Decoder comprises two components: a content part, which corresponds to the decoder output embedding, and a positional part, exemplified by learnable queries in DETR  \cite{detr}. Since the content part of the track query encapsulates semantic information about the referent object  \cite{huang2022minvis}, we employ a momentum-based update strategy to adapt to object feature changes and facilitate long-term tracking. Recent works \cite{defdetr,wang2022anchor,liu2022dabdetr} based on Transformer find that the decoding process in the decoder can be viewed as soft RoI pooling, while anchor boxes provide better spatial priors. Inspired by this, for the position part, we utilize the object bounding box from the last frame to provide a more robust spatial prior \cite{liu2022dabdetr,wang2022anchor}. As validated by the experiment, the proposed query update module improves the accuracy of inter-frame data correlation and simplifies the prediction of reference objects.


In addition, to further advance the development of the referring multi-object tracking task, we introduce the referring multi-object tracking and segmentation (RMOTS). The objective here is to predict the precise spatial-temporal location of objects at the pixel level, which presents a more formidable challenge for RMOTS compared to its bounding box counterpart, RMOT. Besides, object masks, unlike bounding boxes, offer more granular information about the object, thereby leading to broader downstream applications. The proposed dataset, Ref-KITTI Segmentation, is an extension of Ref-KITTI. It consists of 18 videos with 818 expressions, and each expression covers an average of 10.7 instance masks. Notably, this dataset presents a greater challenge compared to existing referring video segmentation datasets, which typically involve only one mask. To further validate the robust cross-modal reasoning capabilities of TenRMOT, we incorporate a segmentation branch upon it. The experimental results demonstrate its superior performance on the newly introduced dataset.


In summary, our work makes the following contributions: 1) We introduce TenRMOT, a model that conducts deep visual-linguistic multi-modal feature fusion at both the encoding and decoding stages. This multi-level feature fusion strategy significantly enhances the accuracy of object predictions as referred to by the language expression. 2) We propose an inter-frame query update module, which explicitly utilizes contextual information and spatial location prior to the referred objects, thus enhancing the temporal consistency of predicted trajectories. 3) To the best of our knowledge, we are the first to introduce the referring multi-object tracking and segmentation task along with the new dataset, Ref-KITTI Segmentation. TenRMOT achieves advanced performance on both the referring multi-object tracking and segmentation tasks.

\section{Related Work}
\label{sec:formatting}

\myPara{Referring Understanding Datasets.}
In the domain of referring understanding, the progression of image-based referent tasks has been substantially facilitated by the introduction of datasets such as ReferIt \cite{kazemzadeh2014referitgame}, Flickr30k \cite{young2014image}, and RefCOCO/+/g \cite{yu2016modeling}. Nevertheless, ambiguous referring language expressions and image-only based shortcomings limit the development of referring tasks in temporal scenes. For this reason, researchers have proposed benchmarks for video scenarios, such as Talk2Car \cite{deruyttere2019talk2car}, VID-Sentence \cite{chen2019weakly}, and CityscapesRef \cite{vasudevan2018object}. In addition, a more fine-grained referring task, referring video object segmentation (RVOS), has also been proposed. Frequently utilized evaluation benchmarks encompass Refer-Youtube-VOS  \cite{seo2020urvos}, Refer-DAVIS17 \cite{khoreva2019video}, and A2D-Sentences \cite{gavrilyuk2018actor}, significantly propelling advancements in video referring tasks. Nonetheless, these benchmarks exhibit two notable limitations, namely a lack of consideration for temporal dynamics changes and the restriction of query expressions to single object only. To this end, the seminal work, RMOT \cite{rmot}, proposes the Refer-KITTI dataset, which can be used as a benchmark for evaluating scenes that refer to multiple objects and have diverse temporal dynamics changes. However, it only uses a 2D bounding box to grounding the target objects, which is somewhat coarse for describing objects, especially in crowded scenes. Pixel-level descriptions, on the other hand, offer greater precision and are more conducive to downstream tasks. To further advance the video referencing task, we propose the Ref-KITTI Segmentation dataset based on the KITTI MOTS \cite{mots} by adding mask annotations to the reference objects in Ref-KITTI.

\begin{figure*}[!t]
  \centering
  \includegraphics[width=0.9\linewidth]{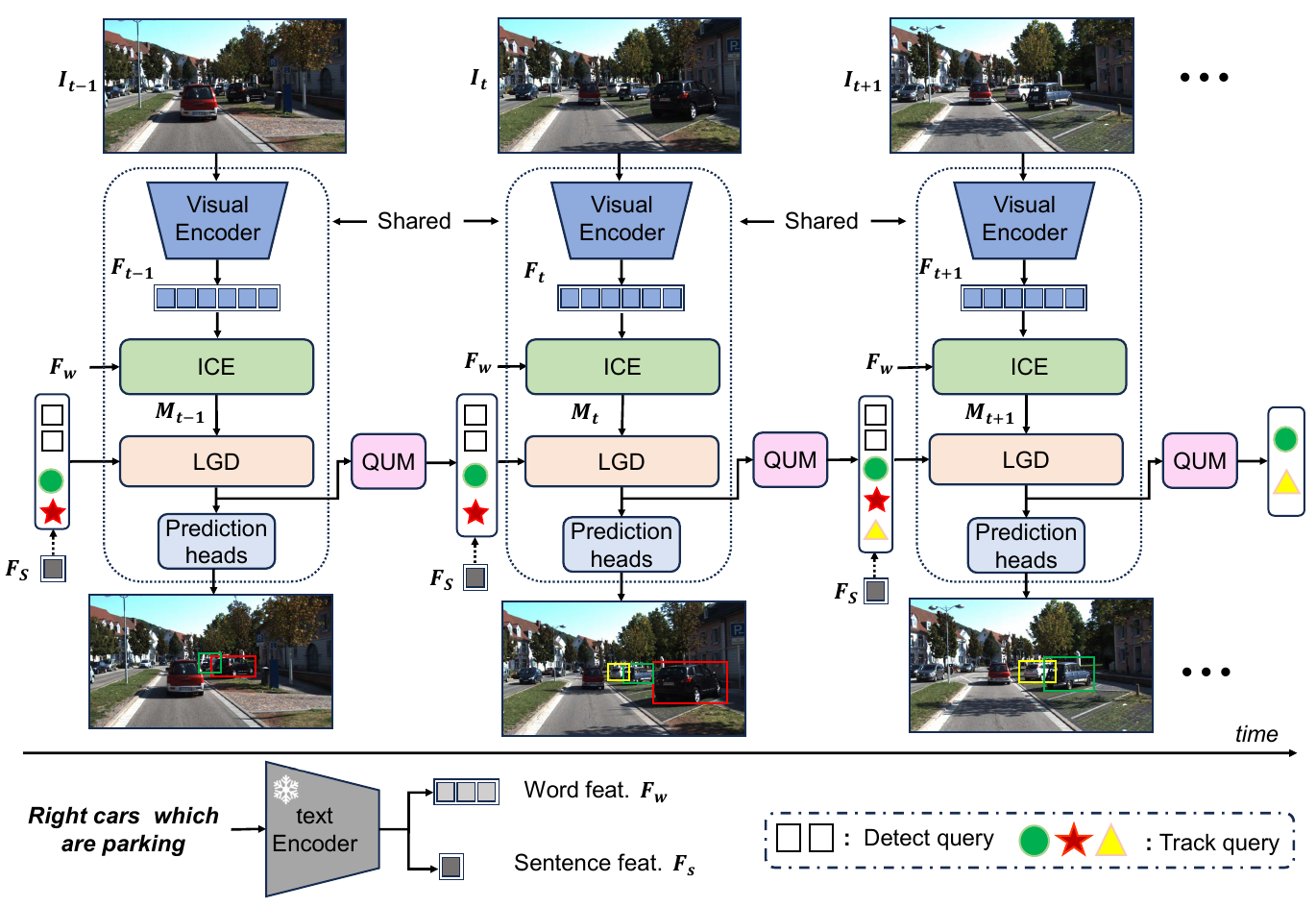}
  \caption{An overview of the proposed TenRMOT. It mainly consists of four parts: feature extraction, Interleaving Cross-modality feature Encoder (ICE), Language Guided Decoder (LGD), and a query updating module (QUM). TenRMOT takes a video sequence $\mathcal{V}$ and a natural language expression $\mathcal{L}$ as input, and outputs the linguistically indicated objects and the corresponding identity labels. TenRMOT conducts vision-language feature fusion at both the decoder-encoder stages, while QUM effectively leverages prior information of tracked objects to update their track queries. Track queries of the same object are identified by sharing the same colored geometric shape. }
  \label{fig:overview}
\end{figure*}

\myPara{Vision-language cross-modal understanding.} Tasks in vision-language (VL) understanding are designed to comprehend the semantic information conveyed in natural language and correlate it with the corresponding visual content. It covers tasks such as referring image/video segmentation \cite{seo2020urvos,kamath2021mdetr,liang2021rethinking, 10463072, 10089870, 9998537}, video captioning \cite{aafaq2019spatio}, image captioning \cite{chen2015microsoft,DingBilateral}, visual navigation \cite{9265290}, etc., and has gained great progress in the recent times. The early approaches \cite{yu2018mattnet, wang2019neighbourhood, nagaraja2016modeling, liu2019learning} employed a two-tower architecture, wherein redundant candidate targets are initially acquired through off-the-shelf detectors. Subsequently, the ultimate referring targets are determined based on the similarity between linguistic expressions and these candidate objects. In recent years, the widespread adoption of Transformer\cite{transformer} structures in visual tasks has led to the application of cross-modal Transformers in the development of one-stage vision-language models\cite{wang2023interrec, zhang2021multi,rmot}. In this study, we meticulously integrate the encoding-decoding architecture of the Transformer with the referring multi-object tracking task, establishing a unified framework for spatial-temporal visual-linguistic grounding.

\myPara{Multi-object tracking.} 
The conventional multi-object tracking task \cite{mot15, mot16,mot20,kitti,dancetrack} aim to localize multiple objects in a video sequence while maintaining their identities. The dominant paradigm \cite{sort,deepsort,centertrack,jde,bytetrack,fairmot, wang2021twostage, XIAO2024106539, feng2020near, kong2022motfr}
utilizes an off-the-shelf detector \cite{yang2019towards} to locate objects, and then links the detections to form a trajectory based on the motion information or appearance feature of the object.

\myPara{End-to-end MOT.}
While significant advancements have been achieved by mainstream tracking algorithms grounded in the tracking-by-detection paradigm, it is noteworthy that the tracking pipeline remains intricate and necessitates meticulous refinement. The pioneering work, DETR \cite{detr}, introduced Transformer to the field of computer vision, enabling end-to-end detection. This inspired a series of end-to-end multi-object algorithms, such as Trackformer \cite{trackformer} and MOTR \cite{motr}. These methods iteratively update the track query of the tracklets frame by frame, and simultaneously achieve object detection and tracking. However, despite their simplicity of structure and elegance of workflow, they have poor detection performance compared to conventional methods. To improve the performance of MOTR, MeMOT \cite{Cai_2022_CVPR} maintains a large memory module and dynamically aggregates tracklet tracking embedding through a memory aggregator module. LTrack \cite{yu2023generalizing} improves the performance of MOTR in unknown domains by introducing a linguistic representation obtained from CLIP \cite{clip}. MOTRv2 \cite{motrv2} uses the detection results of the detector YOLOX \cite{yolox} as anchor boxes to obtain advanced performance on traditional multi-object tracking datasets. Recently, both Co-MOT \cite{comot} and MOTRv3 \cite{motrv3} have found that the cause of poor MOTR detection performance is the unfair label assignment of detect queries and track queries during training. They both propose corresponding improvement strategies to improve the performance of the end-to-end tracker further. Our approach also belongs to the end-to-end tracking methods, but is carefully adapted for the referring multi-object tracking task. In addition, we explicitly utilize the prior information of the tracked object to enhance the inter-frame association and obtain better results.

\section{Method}
\label{sec:method}

Given a video sequence $\mathcal{V} = \{ I_i \}_{i=1}^T$ and a referring expression $\mathcal{L}= \{S_i\}_{i=1}^N$ with $N$ words, our proposed model aims to output the tracks $\mathcal{T}= \{\mathcal{T}_i \}_{i=1}^K$ of the referred $K$ objects and the corresponding identities in an online manner. 
We introduce \textit{TenRMOT}, a new multi-modal transformer architecture constructed upon deformable DETR  \cite{defdetr}. As shown in \cref{fig:overview}, our proposed TenRMOT mainly comprises four components: vanilla visual and text encoders, an Interleaving Cross-modality Encoder (ICE), a Language Guided Decoder (LGD) and a Query Update Module (QUM). We first extract visual and language features from visual and text encoders, respectively. Specifically, we leverage a text encoder to extract the word-level and sentence-level features of $\mathcal{L}$. Besides, a visual encoder is used to extract the feature maps of the video frame $I_t$. $F_w$, $F_s$ and $F_t$ are adjust to the same dimension $d$. Then, $F_w$ and $F_t$ are flattened into sequences and input into the encoder for progressive cross-modality fusion (\cref{sec:encoder}).  In the decoder, the \textit{conditional queries}, which contain sentence-level semantic information of language expression, are used to probe and pool corresponding features from the language-aware memory feature $M$ (\cref{sec:decoder}).
Finally, the output embeddings from the decoder are sent to the prediction heads to predict the desired object boxes. Additionally, we leverage both contextual and spatial priors of the tracked objects in the query update module to improve the consistency of their trajectories (\cref{sec:qum}). TenRMOT shares commonality with previous end-to-end tracking methodologies \cite{trackformer, motr, motrv2, comot} in achieving inter-frame object association through iterative updates of the \textit{track queries} and initialize new tracks with the learned \textit{detect queries}.


\subsection{Interleaving cross-modality encoder}
\label{sec:encoder}

As a visual-linguistic reasoning task, an effective fusion of visual and linguistic features is essential for RMOT to accurately localize the objects referred to in linguistic expressions. 
To achieve this, we introduce a novel interleaving cross-modality encoder named ICE, which consists of  $L$ interleaving encode layers. With the advantage of this layer-by-layer design, the proposed encoder achieves more effective fusion of multi-modal features.
Our proposed encoder takes the vanilla image features and word-level linguistic features as input for cross-modality fusion. 

\begin{figure}[!t]
  \centering
    \includegraphics[width=0.8\linewidth]{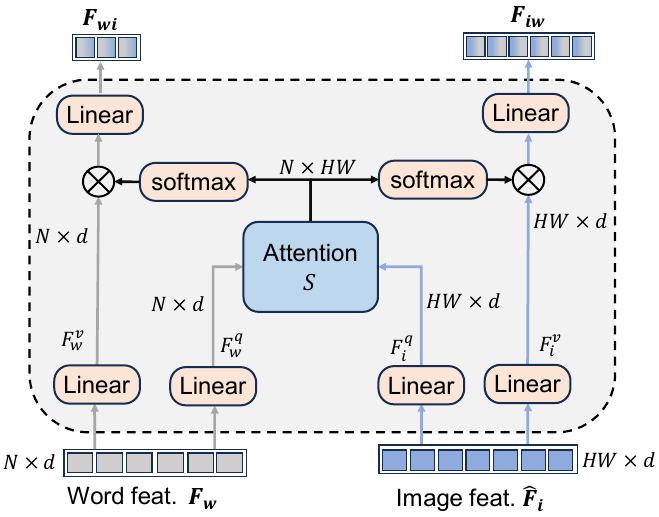}
  \caption{The architecture of the proposed cross-modality attention.}
  \label{fig:attention}
\end{figure}

In each encoding layer, we initially employ deformable self-attention  \cite{defdetr} to acquire the refined visual feature $\hat{F}_i$. 
Inspired by previous research \cite{li2022grounded,tan2019lxmert} on language-image reasoning, we design a bi-directional cross-modality attention module to interleave the enhanced visual feature $\hat{F}_i \in \mathbb{R}^{ HW \times d}$ with the textual features $F_w \in \mathbb{R}^{ N \times d}$. 
As shown in \Cref{fig:attention}, the cross-modality attention can be decomposed into two unidirectional cross-attention operations: one from the vision modality to the language modality, and the other from the language modality to the vision modality:

\begin{equation}
\centering
\small
\begin{aligned}
F_w^q &= F_wW^{(q,L)}, F_i^q = \hat{F}_iW^{(q,I)}, S = F_w^q(F_i^q)^\mathsf{T}/\sqrt{d}, \\
F_i^v &= \hat{F}_iW^{(v,I)}, F_{iw} = \text{softmax}(S)F_i^vW^{(out,I)},  \\
F_w^v &= F_wW^{(v,L)}, F_{wi} = \text{softmax}(S^\mathsf{T})F_w^vW^{(out,L)},
\end{aligned}
\end{equation}
where $S$ is the shared attention map for the two unidirectional cross-attention and $\{ W^{(\text{sym}, L)}, W^{(\text{sym}, I)}, \text{sym} \in \{q, v, out \} \}$ is learned during the training of the model, and acts similarly to the matrix of linear transformations of \textit{query}, \textit{key}, and \textit{value} in the attention mechanism \cite{transformer}. 
Finally, after being processed by ICE, the output $F_{iw}$ of the last encoding layer, i.e., \textit{the encoded memory} $M_t$, is aware of the input language expression. This provides accurate context for the decoding procedure, enabling the model to precisely detect the referring objects.


\subsection{Language guided decoder} 
\label{sec:decoder}

Previous studies \cite{detr, defdetr,wang2022anchor, liu2022dabdetr}, have revealed that the content part and position part of \textit{queries} for decoder contains the content and spatial information respectively. In our language-guided decoder (LGD), we inject the sentence-level features $F_s$ into the content part to guide them to focus specifically on the target objects.

Formally, let $Q_{t-1}^{track} \in \mathbb{R}^{ P \times d} $ denote the content queries of the $P$ active trajectories for the $t-1$ frame and $Q^{det} \in \mathbb{R}^{ N_d \times d} $ denote the content part of the \textit{detect queries} which learned during model training. During the decoding process, the two types of queries are initially concatenated to form $Q \in \mathbb{R}^{ (P + N_d) \times d} $ and then fed into the decoder to probe target objects which correspond to the referring expressions.
In order to make $Q$ only focus on the target objects, we integrate sentence-level features into them to guide the decoding process. 
The sentence features $F_s$ is replicated $P + N_d$ times, added to $Q$ and subsequently passed through the decoder to generate the output content query embedding $E_t \in \mathbb{R}^{ (P + N_d) \times d}$ as follows:

\begin{equation}
\begin{array}{rlrl}
& Q   = \text{concat}(Q^{det}, Q_{t-1}^{track}), \\
& Q_t = \text{repeat}(F_s) + Q, \\
& E_t = \text{Decoder}(M_t, Q_t).
\end{array}
\end{equation}

Consequently, these \textit{conditional queries} $Q_c$ are linguistically guided, focused on locating the referenced objects exclusively. $E_t$ is fed into the subsequent prediction heads to obtain the bounding boxes, masks and confidence scores of the referent objects.

\begin{figure}[!t]
  \centering
    \includegraphics[width=1.0\linewidth]{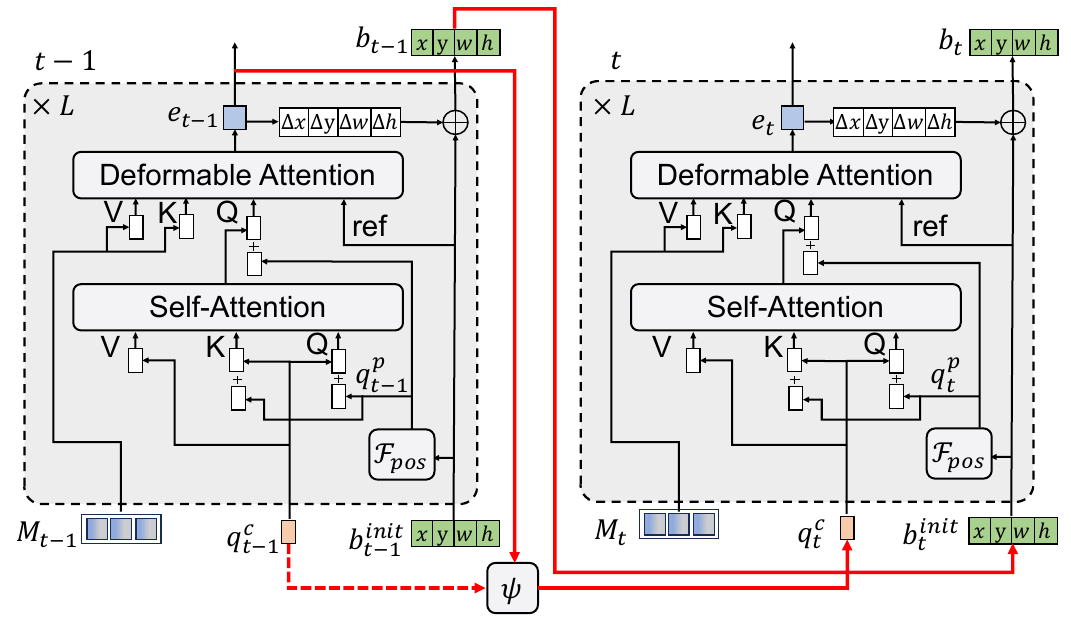}
  \caption{Illustration of inter-frame query update. We explicitly incorporate $q_{t-1}^c$ and $b_{t-1}$ from frame $t -1$ to provide contextual and spatial region prior information for the frame $t$, respectively.}
  \label{fig:query_update}
\end{figure}

\subsection{Inter-frame Query Updating}
\label{sec:qum}
Transformer-based trackers update track queries in an autoregressive manner to track objects in a video. Previous works \cite{motr,rmot} have relied solely on the temporal information from neighboring frames of tracked objects, which may lead to failure in cases of abrupt appearance changes or motion blur.  Inspired by the recent increased understanding of \textit{query} in decoder \cite{huang2022minvis,defdetr,wang2022anchor,liu2022dabdetr}, we propose an inter-frame Query Update Module (QUM) that explicitly leverages both contextual and spatial prior information derived from these tracked objects to improve the association ability.

To be specific, we formulate each query as having two parts: a content part, denoted as $q^c \in \mathbb{R}^{d}$, and a positional part, represented by $q^p \in \mathbb{R}^{4}$ (i.e., the anchor box).
For a tracked object, its content query contains semantic information that provides a contextual prior, and its positional part provides a spatial region prior. As depicted in \cref{fig:query_update}, when updating the track query from frame $t-1$ to frame $t$, we utilize its bounding box $\mathbf{b}_{t-1}$ as initial bounding box $\mathbf{b}_{t}^{init}$ for the current frame. The bounding box $b_t$ for the tracked object in the current frame $t$ is then determined by adding the predicted relative positional offsets to the initial bounding box $b_t^{init}$.
The content query is updated by the momentum update module $\psi(\cdot, \cdot)$ which allows it to incrementally aggregate object features across the entire video, enabling more robust tracking.

Formally, the update procedure is as follows:
\begin{equation}
\begin{aligned}
& \mathbf{b}_{t}^{init}  = \mathbf{b}_{t-1}, \\
& \mathbf{q}_{t}^p  \quad = \mathcal{F}_{pos}(\mathbf{b}_{t}^{init}), \\
& \mathbf{q}_{t}^c  \quad = \psi(\mathbf{e}_{t-1}, \mathbf{q}_{t-1}^c) \\
& \qquad = \alpha \cdot \mathbf{e}_{t-1} + (1 - \alpha) \cdot \mathbf{q}_{t-1}^c,
\end{aligned}
\label{eq:alpha}
\end{equation}

where $\mathcal{F}_{pos}$ refers to one 2-layer feed-forward network (FFN) which generates position embedding. $\alpha$ is a memory update factor to control the update ratio of the content query.

\subsection{Extension to Referring Multi-Object Tracking and Segmentation}
\label{sec: seg_branch}
To further advance the progress in multi-object referring tasks, we introduce the fine-grained Referring Multi-Object Tracking and Segmentation (RMOTS) task along with the Ref-KITTI Segmentation dataset (\cref{sec:exps}).
To equip the proposed TenRMOT model with segmentation capabilities, we extend it by incorporating a segmentation branch.

\begin{figure}[!h]
  \centering
  \includegraphics[width=1.0\linewidth]{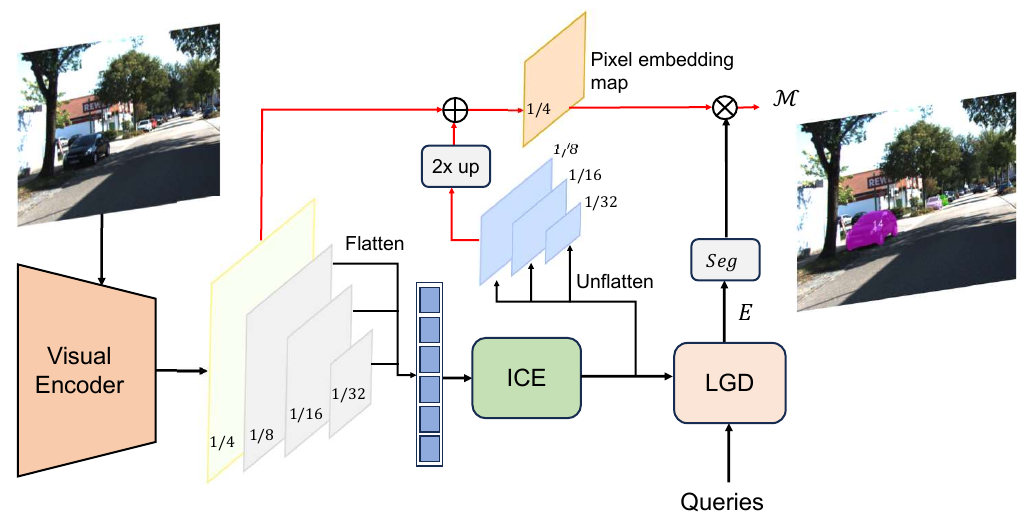}
  \caption{Illustration of the segmentation branch.}
  \label{fig:seg}
\end{figure}

Taking inspiration from Mask2Former \cite{cheng2022masked}, we build a pixel embedding map using the feature maps generated by the visual encoder and Interleaving cross-modality encoder. 
Specifically, as shown in \cref{fig:seg}, the pixel embedding map is constructed by fusing the 1/8 resolution visual-linguistic feature map $\mathbf{F}_{iw}$ from Interleaving cross-modality encoder and 1/4 resolution feature map $f_i$ from the visual encoder. Next, we multiply the output content query embedding $E_t$ from the decoder with the pixel embedding map to obtain the predicted binary mask $\mathcal{M}$:

\begin{equation}
\label{eq:mask}
    \mathcal{M} = seg(E_t)(Conv(f_i) + G(\mathbf{F}_{iw})),
\end{equation}
where $seg$ represents the segmentation head, which consists of a 3-layer feed-forward network (FFN). $Conv$ denotes a convolutional layer used to adjust the channel dimension of $f_i$ to match that of $\mathbf{F}_{iw}$, and $G$ represents a bilinear interpolation function used for up-sampling $\mathbf{F}_{iw}$ to the same resolution as $f_i$. 



\subsection{Instance Matching and Loss}
For the RMOT task, we use the same loss function and label matching cost as TransRMOT\cite{rmot}. For evaluation on the Refer Segmentation, we additionally introduce binary mask focal loss and Dice loss \cite{milletari2016v} to train the mask branch, with corresponding weighting parameters of $\lambda_{\text{mask}}$ and $\lambda_{\text{dice}}$. Furthermore, assuming the model's prediction is $Y$ and the label is $\hat{Y}$, in order to fully utilize the mask label, we use the following matching cost for label assignment:

\begin{equation}
\begin{split}
L_{\text{match}}(Y, \hat{Y}) &= \lambda_{\text{cls}} L_{\text{cls}}(Y, \hat{Y}) + \lambda_{\text{box}} L_{\text{box}}(Y, \hat{Y}) \\
&\quad + \lambda_{\text{mask}} L_{\text{mask}}(Y, \hat{Y})
\end{split}
\label{eq:matching}
\end{equation}


\begin{figure*}[!t]
     \centering
      \subfloat[Distribution of the number of object masks]{\includegraphics[width=0.3\linewidth]
      {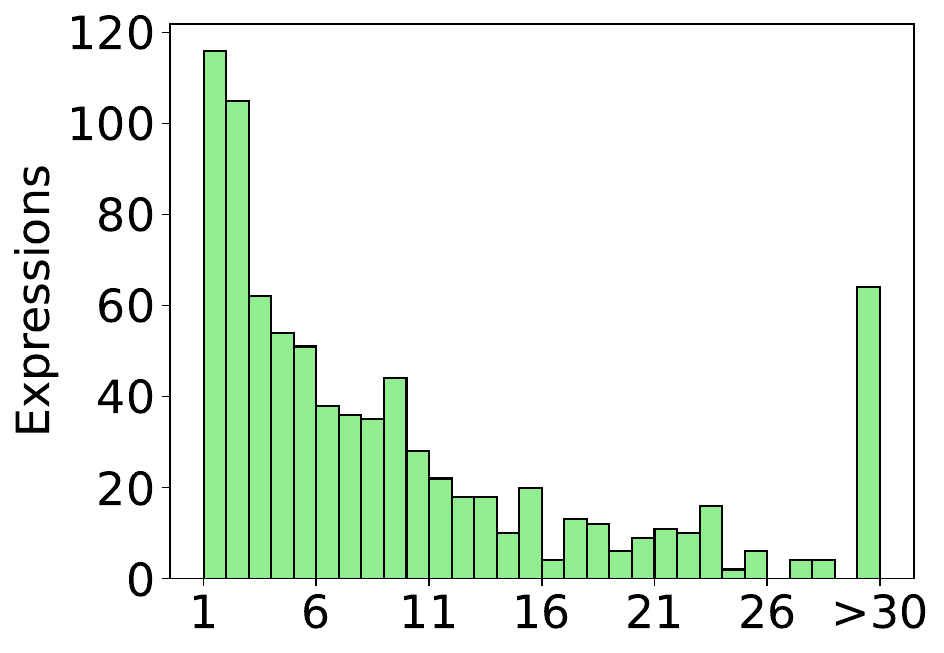}
      
      }
     \hfill
     \subfloat[Distribution of temporal duration per expression]{\includegraphics[width=0.3\linewidth]{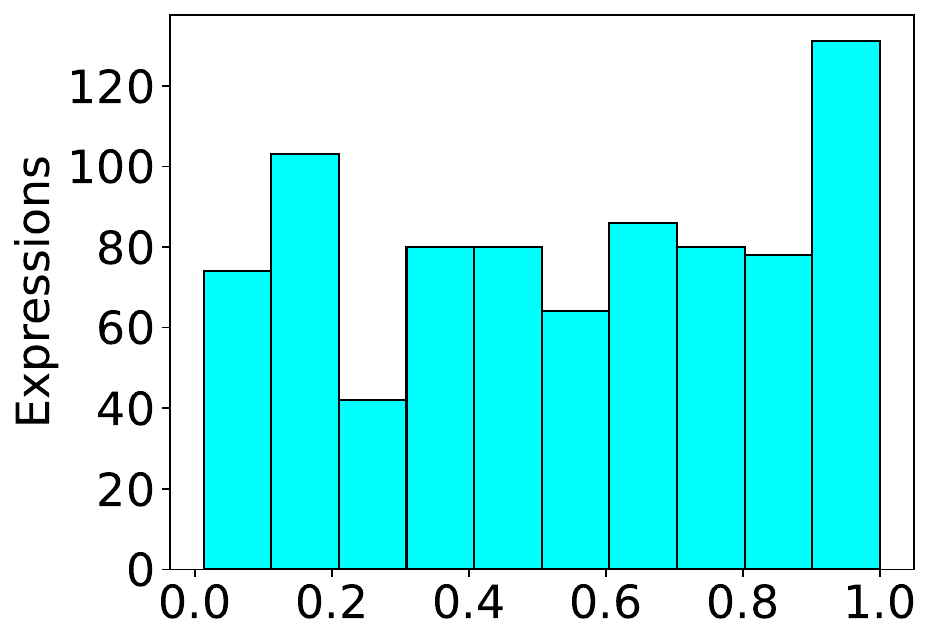}}
     \hfill
     \subfloat[Distribution of frames corresponding to expression]{\includegraphics[width=0.3\linewidth]{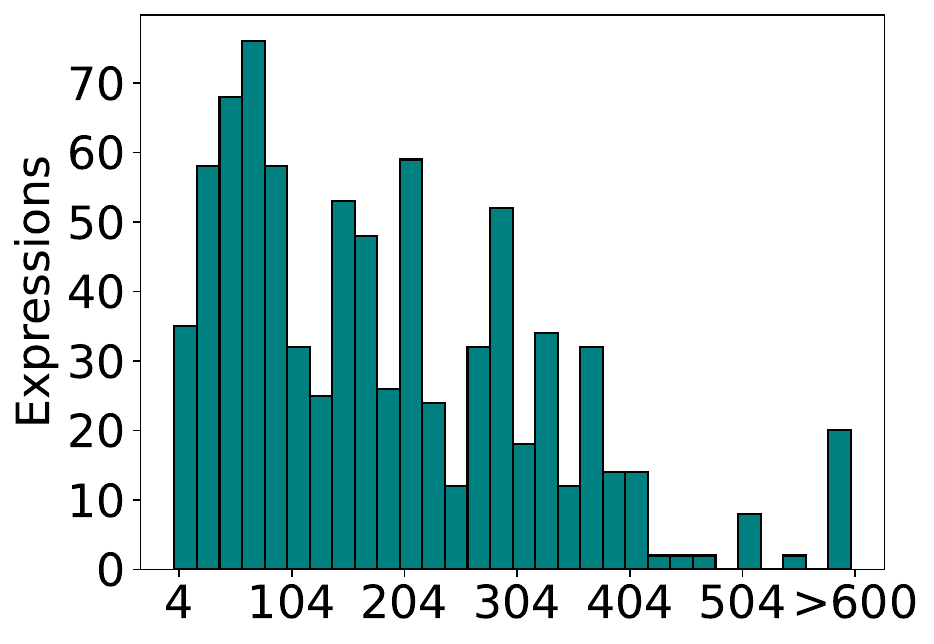}}
\caption{(a) Distribution of the number of object masks. Most language expressions describe 1-26 target objects. (b) Distribution of temporal duration per expressions. This describes the ratio of the video duration during which each referred object exists. (c) Distribution of frames corresponding to expressions. This illustrates the distribution of the number of video frames associated with each language expression. }
\label{fig:rmots_statistics}
\end{figure*}

\begin{figure*}[!t]
  \centering
\includegraphics[width=1.0\linewidth]{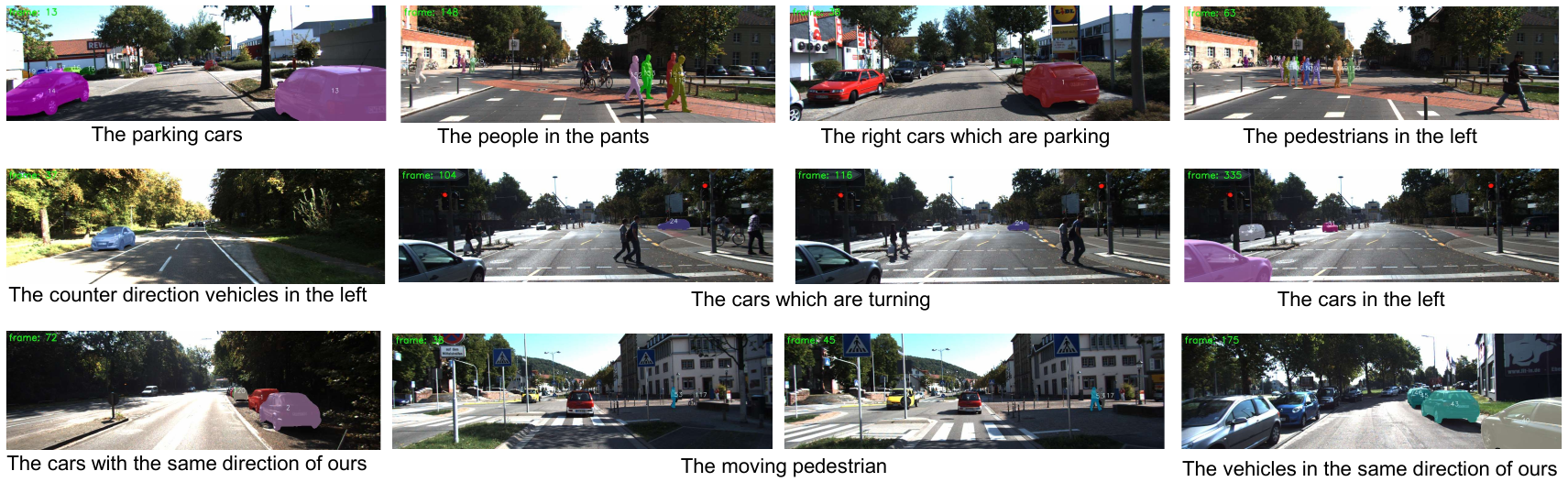}
  \caption{Representative examples from Ref-KITTI Segmentation.}
  \label{fig:example}
\end{figure*}

\section{Experiments}
\label{sec:exps}
\subsection{Datasets and Metrics}

\myPara{Datasets.}  
We evaluate the proposed method on two datasets: Refer-KITTI \cite{rmot}  and Refer-KITTI Segmentation. 
In order to further advance the multi-object referring task, we introduce the Ref-KITTI Segmentation dataset. 
To maximize the use of the existing dataset and minimize human labor cost, we directly extended Ref-KITTI  \cite{rmot}. Specifically, we leveraged the uniqueness of bounding boxes for each target in the video as a bridge, connecting the Ref-KITTI  \cite{rmot} target instances to the target instances in KITTI MOTS\cite{mots}. The proposed dataset contains 18 videos (15 for training and 3 for testing) with 818 expressions, and each expression covers an average of 10.7 objects.
As shown in \cref{fig:rmots_statistics}, the statistics of the dataset in the Ref-KITTI dataset reveal that each language instruction describes multiple target objects, covers videos of various lengths, and each indicated target occupies a diverse proportion of the total video duration.
These statistics show that the newly proposed Ref-KITTI segmentation dataset presents a significant challenge as a benchmark for training and evaluating reference multi-object tracking and segmentation methods, as compared to existing datasets\cite{seo2020urvos,khoreva2019video,gavrilyuk2018actor} that refer to only one object and overlook object temporal variations.
Besides, we illustrate some examples in \cref{fig:example}.

\myPara{Evaluation Metrics.}
Following previous work \cite{rmot}, we employ Higher Order Tracking Accuracy (HOTA) as the primary evaluation metric \cite{hota}. HOTA measures the alignment between the predicted and ground-truth trajectories. It is defined as the geometric mean of detection accuracy (DetA) and association accuracy (AssA), effectively balancing the performance of detection and temporal association. Specifically, HOTA is calculated as \( \text{HOTA} = \sqrt{\text{DetA} \cdot \text{AssA}} \). Additionally, we adopt the following sub-metrics: detection recall/precision (DetRe/DetPr), association recall/precision (AssRe/AssPr), and localization accuracy score (LocA).

When calculating the spatial similarity between the predictions and ground truth, IoU is used for RMOT, while mask IoU is employed for RMOTS. In addition, similar to TransRMOT  \cite{rmot}, if the model predicts these non-referent but visible objects, they are treated as false positives.

\begin{table*}[!t]
\centering
\caption{Comparison to state-of-the-art methods on Refer-KITTI. $\uparrow$ means the higher the better. 
The best results are shown in bold and the second best results are underlined.
}
\begin{tabular}{|l|cccccccc|}
\hline
Method                             & HOTA$\uparrow$ & DetA$\uparrow$ & AssA$\uparrow$ & DetRe$\uparrow$ & DetPr$\uparrow$ & AssRe$\uparrow$ & AssPr$\uparrow$ & LocA$\uparrow$ \\ \hline
\textit{CNN-based:}                & & & & & & & & \\
FairMOT \cite{fairmot}             & 22.78          & 14.43          & 39.11          & 16.44           & 45.48           & 43.05           & 71.65           & 74.77          \\
DeepSORT \cite{deepsort}           & 25.59          & 19.76          & 34.31          & 26.38           & 36.93           & 39.55           & 61.05           & 71.34          \\
ByteTrack \cite{bytetrack}         & 24.95          & 15.50          & 43.11          & 18.25           & 43.48           & 48.64           & 70.72           & 73.90          \\
CStrack \cite{liang2022rethinking} & 27.91          & 20.65          & 39.10          & 33.76           & 32.61           & 43.12           & 71.82           & 79.51          \\
iKUN \cite{du2024ikun}             & {\ul 48.84}    & 35.74          & \textbf{66.80} & {\ul 51.97}     & 52.25           & \textbf{72.95}  & 87.09           & -              \\
\hline

\textit{Transformer-based:}                & & & & & & & & \\
TransTrack \cite{transtrack}       & 32.77          & 23.31          & 45.71          & 32.33           & 42.23           & 49.99           & 78.74           & 79.48          \\
DeepRMOT \cite{he2024visual}       & 39.55          & 30.12          & 53.23          & 41.91           & 47.47           & 58.47           & 82.16           & 80.49          \\
TrackFormer \cite{trackformer}     & 33.26          & 25.44          & 45.87          & 35.21           & 42.19           & 50.26           & 78.92           & 79.63          \\
TransRMOT \cite{rmot}              & 46.56          & {\ul 37.97}    & 57.33          & 49.69           & {\ul 60.10}     & 60.02           & \textbf{89.67}  & {\ul 90.33}    \\
TenRMOT (Ours)                     & \textbf{49.77} & \textbf{40.79} & {\ul 60.89}    & \textbf{52.65}  & \textbf{62.81}  & {\ul 65.38}     & {\ul 89.28}     & \textbf{90.69} \\ \hline
\end{tabular}
\label{table:sota_kitti}
\end{table*}

\begin{table*}[!t]
\centering
\caption{Quantitative results on Refer-KITTI Segmentation.  $\uparrow$ means the higher the better. 
The best results are shown in bold and the second best results are underlined. 
}
\begin{tabular}{|l|cccccccc|}
\hline
Method                                & HOTA$\uparrow$ & DetA$\uparrow$ & AssA$\uparrow$ & DetRe$\uparrow$ & DetPr$\uparrow$ & AssRe$\uparrow$ & AssPr$\uparrow$ & LocA$\uparrow$ \\ \hline
\textit{CNN-based:}                & & & & & & & & \\
Mask-Track R-CNN \cite{yang2019video} & 13.67          & 10.47          & 18.02          & 28.80           & 14.00           & 49.40           & 22.86           & 87.23          \\
QDTrack \cite{fischer2023qdtrack}     & 37.73          & 23.81          & \textbf{60.04} & \textbf{65.79}  & 26.66           & \textbf{70.37}  & 77.78           & \textbf{87.48} \\ \hline
\textit{Transformer-based:}                & & & & & & & & \\
TrackFormer \cite{trackformer}        & 30.95          & 23.20          & 42.07          & 32.64           & 42.03           & 48.40           & 64.85           & 82.28          \\
TransRMOT \cite{rmot}                 & {\ul 44.17}    & {\ul 35.63}          & 55.21          & {\ul 50.39}     & {\ul 52.66}     & 60.01           & {\ul 86.44}     & {\ul 87.21}    \\
TenRMOT (Ours)                        & \textbf{46.00} & \textbf{36.68} & {\ul 58.17}    & 49.70           & \textbf{55.81}  & {\ul 62.31}     & \textbf{87.44}  & 87.12          \\ \hline
\end{tabular}
\label{table:sota_kitti_mots}
\end{table*}

\subsection{Experimental Setting}

\myPara{Model Settings.} 
We leverage ResNet-50   \cite{he2016deep} as visual encoder and RoBERTa  \cite{roberta} as text encoder, separately. As with deformable DETR  \cite{defdetr}, we adopt the vision features from the last three stage of the visual backbone as the input to Transformer. For the Transformer component, we utilize 6 encode layers and 6 decode layers. The number $N_d$ of \textit{detect query} is set as 300. Because of memory limitation of GPU, $N_d$ is set as 200 for segmentation.

\myPara{Training Details.} 
The learnable parameters associated with multi-modal feature fusion are initialized with random values, and the parameters of RoBERTa remain frozen during training. For both datasets, we initialize the remaining parameters of TenRMOT with official pre-trained weights from deformable DETR  \cite{defdetr} on the COCO  \cite{chen2015microsoft} dataset. The same data augmentation strategies as those in TransRMOT are applied. The proposed model is trained using the AdamW optimizer, with an initial learning rate of $1e^{-5}$ for the visual backbone and $1e^{-4}$ for the Transformer part. The weighting coefficients for the loss functions mostly follow TransRMOT: $\lambda_{cls} = 5$, $\lambda_{giou} = 2$, $\lambda_{L_1} = 2$, $\lambda_{ref} = 2$. For the segmentation dataset, we add mask loss and dice loss follow DETR \cite{detr}, $\lambda_{mask} = 5$ and $\lambda_{dice} = 5$. All experiments were conducted using eight Tesla V100 GPUs, with a batch size of 1 on each GPU. We train TenRMOT for 60 epochs and the learning rate decays by 10 at the $50^{th}$ epoch.

\myPara{Inference.}
As an  end-to-end multi-object tracking algorithm, TenRMOT eliminates the need for post-processing of prediction results, such as non-maximum suppression (NMS). Following TransMOT  \cite{rmot}, the prediction results for the current frame are filtered using a confidence threshold $\beta_{conf} = 0.7 $ and a referring threshold $\beta_{ref} = 0.4$ pair.

\subsection{Comparison With State-of-the-Art Methods}

\myPara{Ref-KITTI.} As shown in \Cref{table:sota_kitti}, it is clear that TenRMOT outperforms existing methods which have been officially published across several evaluation metrics. On the primary evaluation metric, HOTA, our approach outperforms all CNN-based models by a substantial margin. Specially, our method outperforms iKUN \cite{du2024ikun} exploiting the foundation model CLIP \cite{clip}. The proposed method achieves a HOTA score of 49.77\% and demonstrates a substantial lead on the DetA metric. Besides, TenRMOT surpasses the previous Transformer-based referring multi-object tracking model, TransRMOT \cite{rmot}, by impressive margins of \textbf{2.82} and \textbf{3.56} percentage points on the DetA and AssA metrics, respectively. This outstanding performance in detection and association is attributed to the effectiveness of our proposed compact cross-modal feature fusion, facilitating the efficient fusion of visual and linguistic features, coupled with the comprehensive utilization of temporal prior information of the referenced objects. 

\myPara{Ref-KITTI Segmentation.}
We further evaluate TenRMOT on the newly introduced and challenging Ref-KITTI Segmentation dataset, as shown in \Cref{table:sota_kitti_mots}. We follow RMOT \cite{rmot} to extend existing multi-object tracking and segmentation models \cite{yang2019video,fischer2023qdtrack,trackformer} with a cross-modal fusion component to achieve RMOTS task. Additionally, since TransMOT lacks segmentation capabilities, we incorporate the same segmentation branch proposed in this paper to ensure fairness in the comparison (\cref{sec: seg_branch}). As depicted in \cref{table:sota_kitti_mots}, TenRMOT exhibits a performance advantage over the baseline method, TransRMOT, by \textbf{1.83}\% in the HOTA metric, providing further validation of its effectiveness.

\subsection{Qualitative Results}
\begin{figure*}[!t]
  \centering
  \includegraphics[width=1.0\linewidth]{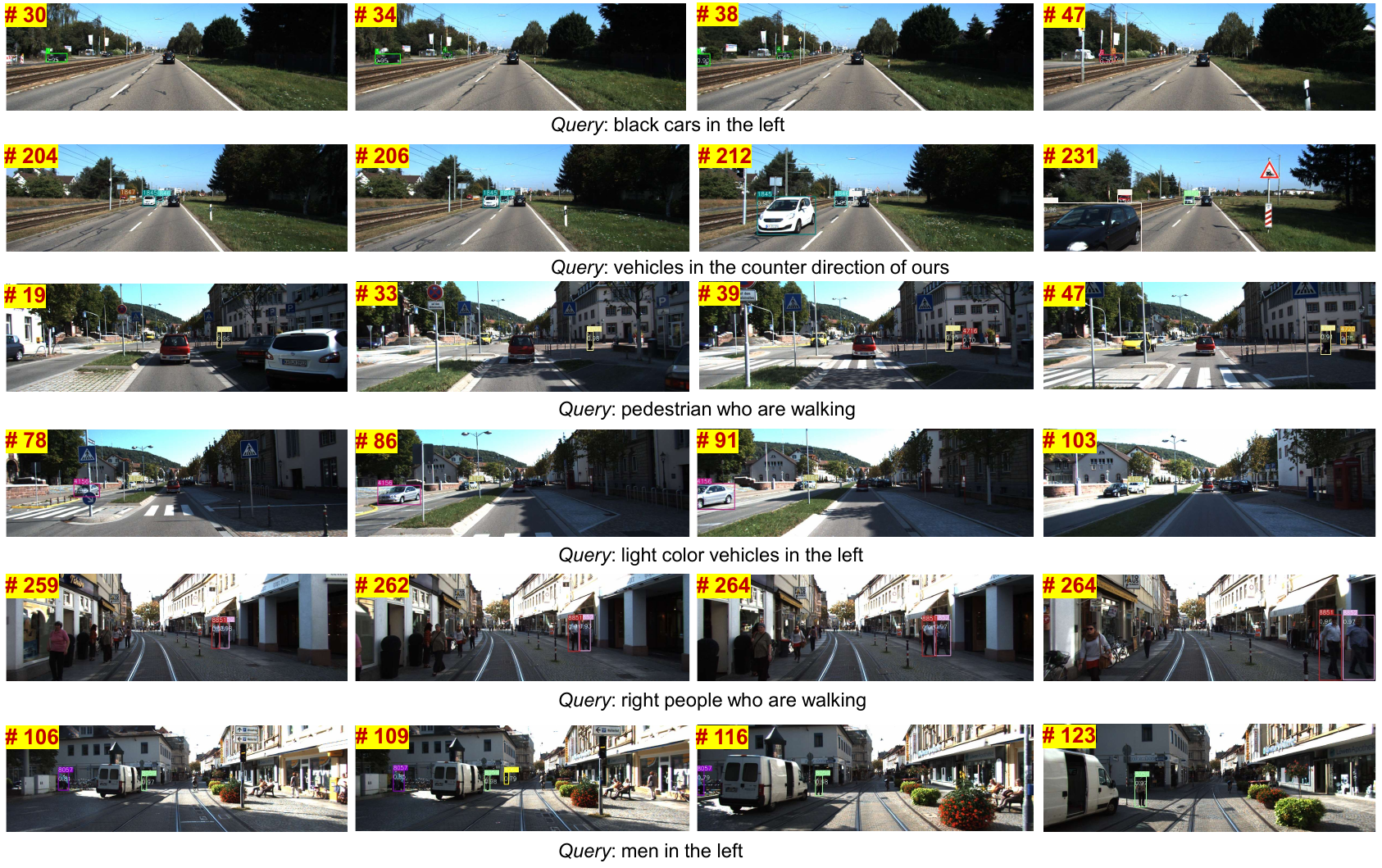}
  \centering
  \caption{Qualitative results on Refer-KITTI. The different colors of the bounding boxes in the figure represent the identities of different objects. Kindly magnify the figures to observe finer details.}
  \label{fig:qualitative}
\end{figure*}

\begin{figure*}[!h]
  \centering
  \includegraphics[width=1.0\linewidth]{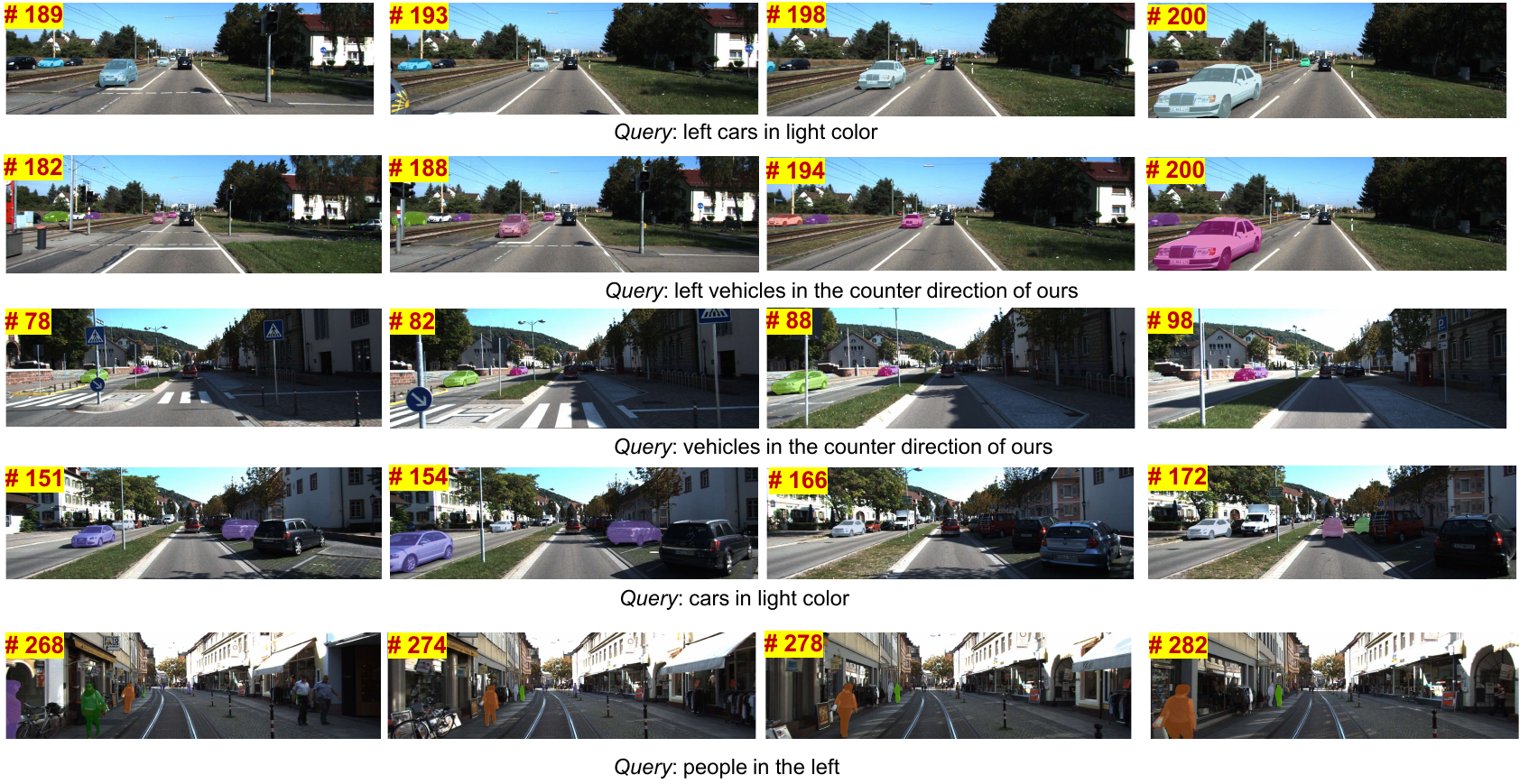}
  \caption{Qualitative results on Refer-KITTI Segmentation. The different colors of the masks in the figure represent the identities of different objects. Kindly magnify the figures to observe finer details.}
  \label{fig:qualitative2}
\end{figure*}

We present the quantitative results of TenRMOT on Refer-KITTI and Refer-KITTI Segmentation datasets in \Cref{fig:qualitative} and \Cref{fig:qualitative2}, separately. As depicted in the figure, our proposed TenRMOT demonstrates accurate detection and tracking of objects specified by query expression even in challenging scenarios that involve multiple indicated objects, changes in the state of indicated objects, and variations in the number of indicated objects.

\subsection{Ablation Study}

\begin{figure}[!h]
  \centering
  \includegraphics[width=1.0\linewidth]{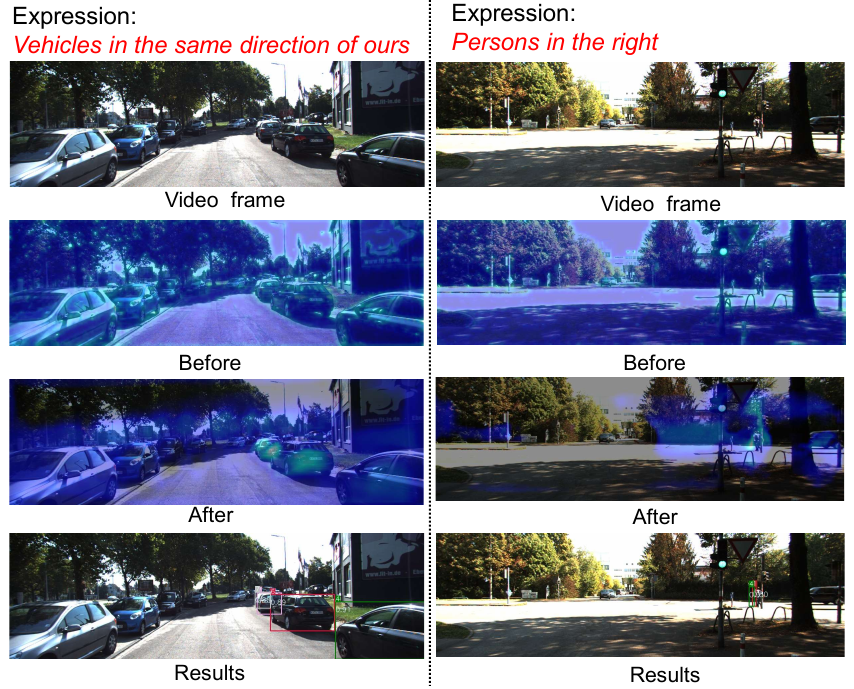}
  \caption{Visualization of the visual features before and after the proposed interleaving cross-modality encoder.}
  \label{fig:feature_vis}
\end{figure}

\myPara{Effectiveness of the model components. }
We first conduct ablation experiments to study and validate the effectiveness of the core designs in TenRMOT.  The components are built upon the pioneering work, TransRMOT \cite{rmot}, step by step. As observed in \Cref{table: comp.}, each of the proposed design modifications exhibits a positive impact on the overall performance. The utilization of interleaving cross-modality encoder for cross-modality feature fusion markedly enhances model performance. This highlights that our proposed encoder, composed of interleaving coding layers, achieves more effective feature fusion, fully exploiting the advantages of the Transformer layer-by-layer structure. As shown in \Cref{fig:feature_vis}, the proposed interleaving cross-modality encoder effectively disentangles objects from other objects and background, highlighting the objects referenced by the query language expression. Moreover, leveraging sentence-level semantic information at the decoder side to guide the decoding process yields positive gains. The incorporation of the query update module further enhances model performance. 
When ICE and LGD were individually removed from TenRMOT in experiments \textcircled{5} and \textcircled{6}, respectively, the model's performance declined in both cases. Specifically, we removed the ICE module from TenRMOT and retained only LGD for linguistic semantic guidance (experiment \textcircled{5}). As shown in \Cref{table: comp.}, the model performance drops dramatically, with a HOTA of only 11.72\%. This suggests that visual and linguistic feature fusion at the encoding stage is crucial.

\begin{table}[!t]
\centering
\setlength{\tabcolsep}{1.4 mm}
\caption{Ablation of different components on TenRMOT}
\begin{tabular}{|c|ccc|ccccc|}
\hline
                & ICE          & LGD          & QUM          & HOTA$\uparrow$ & DetA$\uparrow$ & AssA$\uparrow$ & DetRe$\uparrow$ & DetPr$\uparrow$         \\ \hline
\textcircled{1} &              &              &              & 46.56          & 37.97          & 57.33          & 49.69          & 60.10         \\
\textcircled{2} & $\checkmark$ &              &              & 47.52          & 38.29          & 59.13          & 47.36          & \textbf{65.04}         \\
\textcircled{3} & $\checkmark$ & $\checkmark$ &              & 48.42          & 39.40          & 59.63          & 50.98          & 61.86          \\
\textcircled{4} & $\checkmark$ & $\checkmark$ & $\checkmark$ & \textbf{49.77} & \textbf{40.79} & \textbf{60.89} & \textbf{52.65} & 62.81          \\ 
\textcircled{5} &              & $\checkmark$ & $\checkmark$ & 11.72          & 6.40           & 21.56          & 7.91           & 24.86          \\
\textcircled{6} & $\checkmark$ &             & $\checkmark$  & 47.74          & 39.57          & 57.76          & 51.92          & 60.86          \\
\hline
\end{tabular}
\label{table: comp.}
\end{table}

\myPara{Effect of strategies for using linguistic features in the decoder}.
In \Cref{table: lgd_design}, we investigate the influence of various language feature utilization strategies in the Language Guided Decoder (LGD) on the model's performance, specifically considering the injection of language features into both detection and tracking queries. The findings reveal that TenRMOT attains a HOTA score of 48.80\% when no linguistic features are incorporated into the decoder. Nevertheless, introducing linguistic features into detect and track queries results in performance enhancements, suggesting that the inclusion of linguistic semantic features in the decoder aids TenRMOT in more accurately localizing the target objects specified by language expressions.

\begin{table}[!t]
\centering
\setlength{\tabcolsep}{1.2mm}
\caption{Effect of strategies for using linguistic features in the decoder}
\begin{tabular}{|cc|ccccc|}
\hline
\textit{detect queries} &\textit{track queries}  & HOTA$\uparrow$ & DetA$\uparrow$ & AssA$\uparrow$ & DetRe$\uparrow$ & DetPr$\uparrow$    \\ 
\hline
&                              & 48.80          & 39.90 & 59.83  & \textbf{55.53}   & 57.19    \\
$\checkmark$ &                 & 48.84          & 40.09 & 59.66 & 51.41 & \textbf{62.91}                             \\
$\checkmark$ &$\checkmark$     & \textbf{49.77} & \textbf{40.79} & \textbf{60.89} & 52.65 & 62.81    \\
\hline
\end{tabular}
\label{table: lgd_design}
\end{table}

\myPara{Effect of  Query Update Module (QUM).}
As detailed in the main text (\Cref{sec:qum}), the track queries of objects comprise a content part and a positional part. The effectiveness of the proposed design is validated in \Cref{table: qum_design}. When our proposed query update strategy is not employed, i.e., using the query update strategy introduced by MOTR\cite{motr}, a HOTA score of 48.42\% is obtained. Leveraging the anchor boxes instead of the reference points used by MOTR to provide spatial prior results in a 1.21 percentage point improvement in HOTA. The model performance is further improved when the content part of the tracking query is further updated in a momentum manner. Consequently, the experimental results in \Cref{table: qum_design} affirm the effectiveness of the proposed Query Update Module (QUM).

\begin{figure*}[!t]
     \centering
      \subfloat[The referring expression "turning car" spans multiple video frames.]{\includegraphics[width=1.0\linewidth]
      {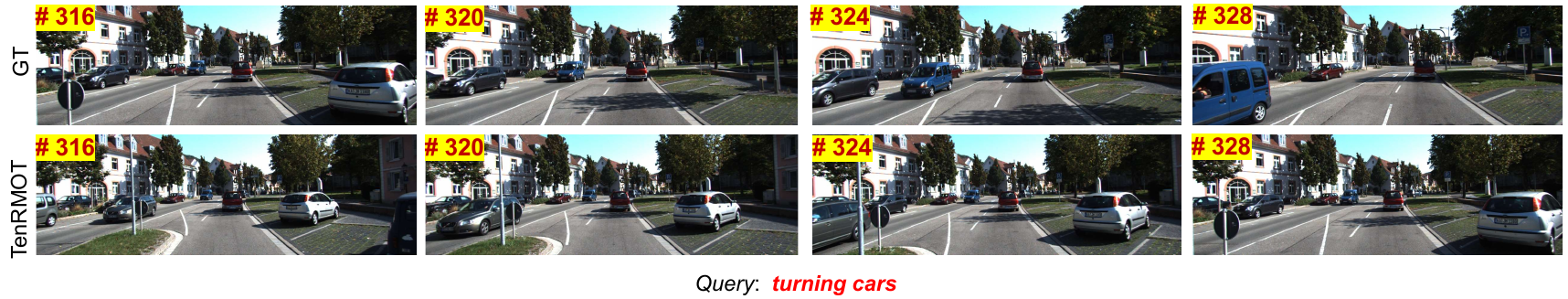}
      }
     \vfill
     \subfloat[The car parked on the left side of frame 104 is moving relative to the camera's perspective.]{\includegraphics[width=1.0\linewidth]{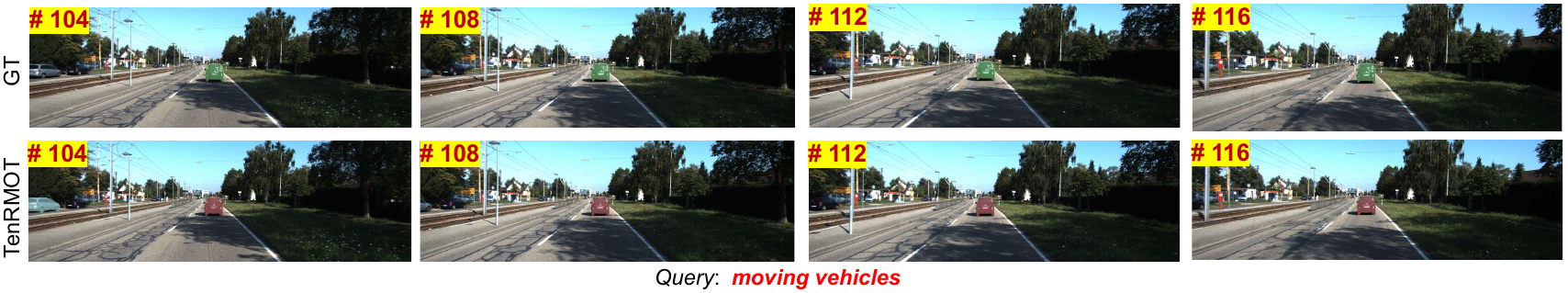}}
\caption{Instances of failure of the proposed TenRMOT on Ref-KITTI segmentation dataset.}
\label{fig:failures}
\end{figure*}

\begin{table}[!t]
\centering
\setlength{\tabcolsep}{0.8mm}
\caption{Effect of Query Update Module}
\begin{tabular}{|cc|ccccc|}
\hline
anchor boxes   & content update  & HOTA$\uparrow$ & DetA$\uparrow$ & AssA$\uparrow$ & DetRe$\uparrow$ & DetPr$\uparrow$           \\ \hline
               &              & 48.42          & 39.40          & 59.63          & 50.98          & 61.86          \\
$\checkmark$   &              & 49.63          & \textbf{40.85}          & 60.52          & \textbf{55.71}          & 58.86  \\
$\checkmark$ &$\checkmark$     & \textbf{49.77} & 40.79 & \textbf{60.89} & 52.65 & 62.81 \\
\hline
\end{tabular}
\label{table: qum_design}
\end{table}

\begin{figure}[!t]
  \centering
  \includegraphics[width=1.0\linewidth]{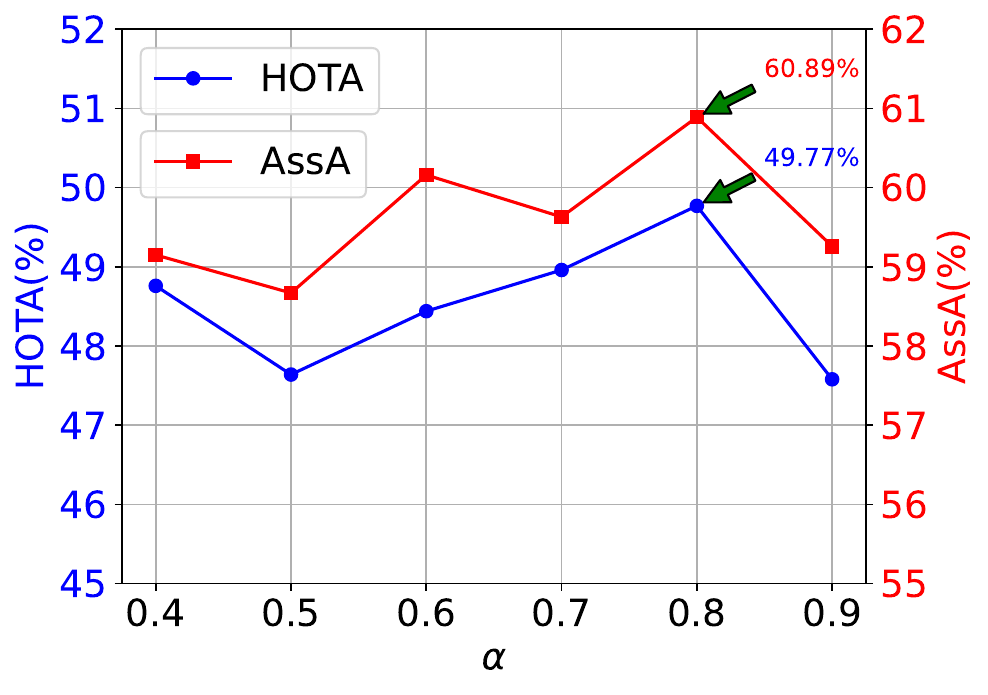}
  \caption{Effect of different content query update factor $\alpha$.}
  \label{fig:alpha}
\end{figure}

\myPara{Effect of different content query update factor.}
We investigate the impact of different factor $\alpha$ in \Cref{eq:alpha} on the performance of TenRMOT further. As shown in \Cref{fig:alpha}, the best performance is achieved when $\alpha=0.8$, with HOTA and AssA reaching 49.77\% and 60.89\%, respectively. 

\myPara{Analysis of inference time consumption.} We tested the inference elapsed time of our proposed TenRMOT and the baseline work TransRMOT on the same device (a laptop with an RTX 4060). The "First frame" represents the processing elapsed time of the first frame in milliseconds, and the processing speed of the video sequence is measured in frames per second (FPS). As shown in \Cref{table: infer_time}, our method outperforms TransRMOT in both the processing speed of the first frame and the processing speed of the whole video sequence. Considering that in the main evaluation metric, HOTA, our method significantly outperforms TransRMOT, paying a little extra inference time cost is acceptable.

\begin{figure}[!t]
  \centering
  \includegraphics[width=1.0\linewidth]{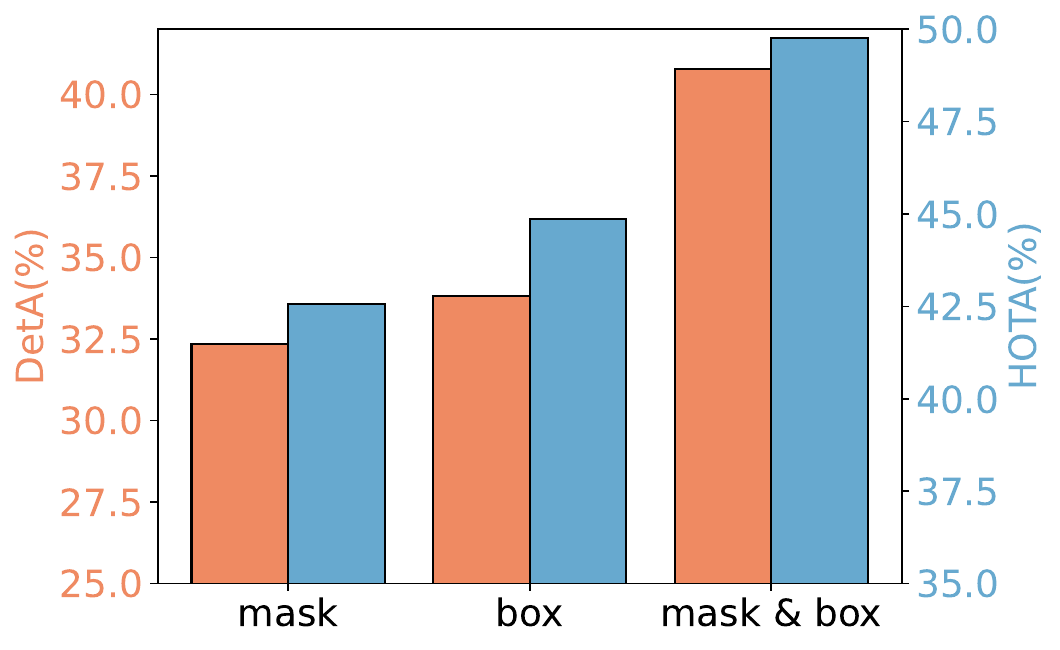}
  \caption{Effect of different label matching cues.}
  \label{fig:matching_cues}
\end{figure}
\myPara{Effect of different label matching cues.}
Compared to the Ref-KITTI dataset, the Ref-KITTI segmentation dataset provides mask annotations, offering more fine-grained annotation information that facilitates more accurate label assignment during training for RMOTS task.
As illustrated in \Cref{fig:matching_cues}, we assess the utilization of various cues to compute the matching cost in \Cref{eq:matching}. It is evident from this table that the lack of either the mask or box results in a degradation in performance. Yet, when both are integrated, a more robust label assignment is achieved, which improves the guidance for model training.

\begin{table}[!t]
\centering
\setlength{\tabcolsep}{2.4mm}
\caption{Comparison of inference time consumption}
\begin{tabular}{|l|c|c|}
\hline
Method & First frame (ms) & Video sequence (FPS) \\
\hline
TransRMOT                  & 883              & 4.32                 \\
TenRMOT(Ours)              & 919              & 4.80             \\   
\hline
\end{tabular}
\label{table: infer_time}
\end{table}

\subsection{Limitations}
Despite the advancements achieved by the proposed TenRMOT, its online nature (limited to processing video streams frame by frame) and the intricacies associated with the RMOTS contribute to the unsatisfactory performance of our method in some scenarios.
As shown in \Cref{fig:failures}, we demonstrate two typical failure cases on Ref-KITTI segmentation dataset: 
(1) instance of query expression spanning multiple video frames.
As illustrated in \Cref{fig:failures} (a), the query language "turning car" spans multiple video frames, making it challenging to track the referred objects solely relying on single frame information.
(2) Instance of unclear referring expression. As shown in \Cref{fig:failures} (b), regarding the expression "moving vehicle", the car parked on the left side of the road in frame 104 of the ground truth is not the referenced object. However, from the camera's perspective, it is in motion. Consequently, our method highlights it as the target object. In such cases, the query expression is ambiguous, and the model struggles to understand. Exploring strategies to address such challenging scenarios in future research would be an intriguing research topic. In future work, we will explore the utilization of multi-frame information to address such temporal-wise challenges.

\section{Conclusion}
In this paper, we present TenRMOT, a compact framework designed for referring multi-object tracking. TenRMOT performs vision-language feature fusion at both the encoding and decoding stages, enhancing the localization of the objects referenced by query expression. Additionally, we explicitly utilize contextual and spatial region information of tracked objects to enhance robustness against occlusion and irregular motion. Furthermore, we introduce a new benchmark, Refer-KITTI Segmentation, which provides a more challenging setting for referring video segmentation. Our model is evaluated on Ref-KITTI and Ref-KITTI segmentation, demonstrating superior performance on both benchmarks. We wish that our proposed method and benchmark will promote further research in the domain of referring dynamic object tracking and video understanding.

\small
\bibliographystyle{IEEEtran}
\bibliography{ref.bib}


 




\vfill

\end{document}